# VORTEX: A Spatial Computing Framework for Optimized Drone Telemetry Extraction from First-Person View Flight Data

**JAMES E GALLAGHER[1], EDWARD J OUGHTON[1]**

[1] Geography & Geoinformation Science Department, George Mason University, Fairfax, VA 22030, USA.

Corresponding author: James E. Gallagher (e-mail: jgalla5@gmu.edu).

This work was supported by the Geography & Geoinformation Science Department at George Mason University.

**ABSTRACT** This paper presents the Visual Optical Recognition Telemetry EXtraction (VORTEX) system for analyzing drone telemetry data from First Person View (FPV) Uncrewed Aerial System (UAS) footage. VORTEX employs MMOCR, a PyTorch-based Optical Character Recognition (OCR) toolbox, to extract telemetry variables from drone Heads Up Display (HUD) recordings, utilizing advanced image preprocessing techniques, including Contrast Limited Adaptive Histogram Equalization (CLAHE) enhancement and adaptive thresholding. The study optimizes spatial accuracy and computational efficiency through systematic investigation of temporal sampling rates (1s, 5s, 10s, 15s, 20s) and coordinate processing methods. Results demonstrate that the 5-second sampling rate, utilizing 4% of available frames, provides the optimal balance with a point retention rate of 64% and mean speed accuracy within 4.2% of the 1-second baseline, while reducing computational overhead by 80.5%. Comparative analysis of coordinate processing methods reveals that while UTM Zone 33N projection and Haversine calculations provide consistently similar results (within 0.1% difference), raw WGS84 coordinates underestimate distances by 15-30% and speeds by 20-35%. Altitude measurements showed unexpected resilience to sampling rate variations, with only 2.1% variation across all intervals. This research provides unique quantitative benchmarks for establishing a robust framework for drone telemetry extraction and analysis using open-source tools and spatial libraries.

## I. INTRODUCTION

The use of Uncrewed Aerial Systems (UAS) is proliferating in all sectors of society, from simple drone hobbyists and commercial applications to UAS in warfare [1]. Given their effectiveness and low cost, UAS is only projected to grow in use [2], [3]. Due to this continual growth, there needs to be open-source analytical tools that enable drone operators to analyze critical flight data [4]. Current methods to extract raw telemetry from drones are challenging, especially when using proprietary software and hardware that does not allow data export outside the manufacturer's ecosystem [5].

A computer-vision telemetry extraction solution is optimal for UAS flight analysis when telemetry data is lost from the flight controller or the UAS loses link with the ground-control station. Using telemetry extraction and mapping, drone operators can use recorded flight video from the ground station or First Person View (FPV) goggles to analyze flight paths and other telemetry data before losing the link to the aircraft, helping determine the cause of incidents and project probable crash sites.

The Visual Optical Recognition Telemetry EXtraction (VORTEX) System was developed to extract and analyze drone-derived data from the Heads Up Display (HUD). Figure 1 visualizes how VORTEX conducts Optical Character Recognition (OCR) on data derived from an FPV HUD screen. During drone flights, an FPV pilot may want to analyze flight data variables, such as flight path, altitude, air speed, and other telemetry data. However, extracting telemetry data from a non-commercial flight controller can prove challenging. Since HUD data is recorded from the FPV pilot's goggles, VORTEX can easily extract relevant data and visualize the results by exporting the HUD variables as .csv and .kmz files, allowing for data processing without infiltrating proprietary programs. The processed .csv data can analyze variables such as battery life, airspeed, altitude, direction, location, and many other variables. At the same time, the .kmz file can visualize location data (latitude, longitude, and altitude) through GIS programs.

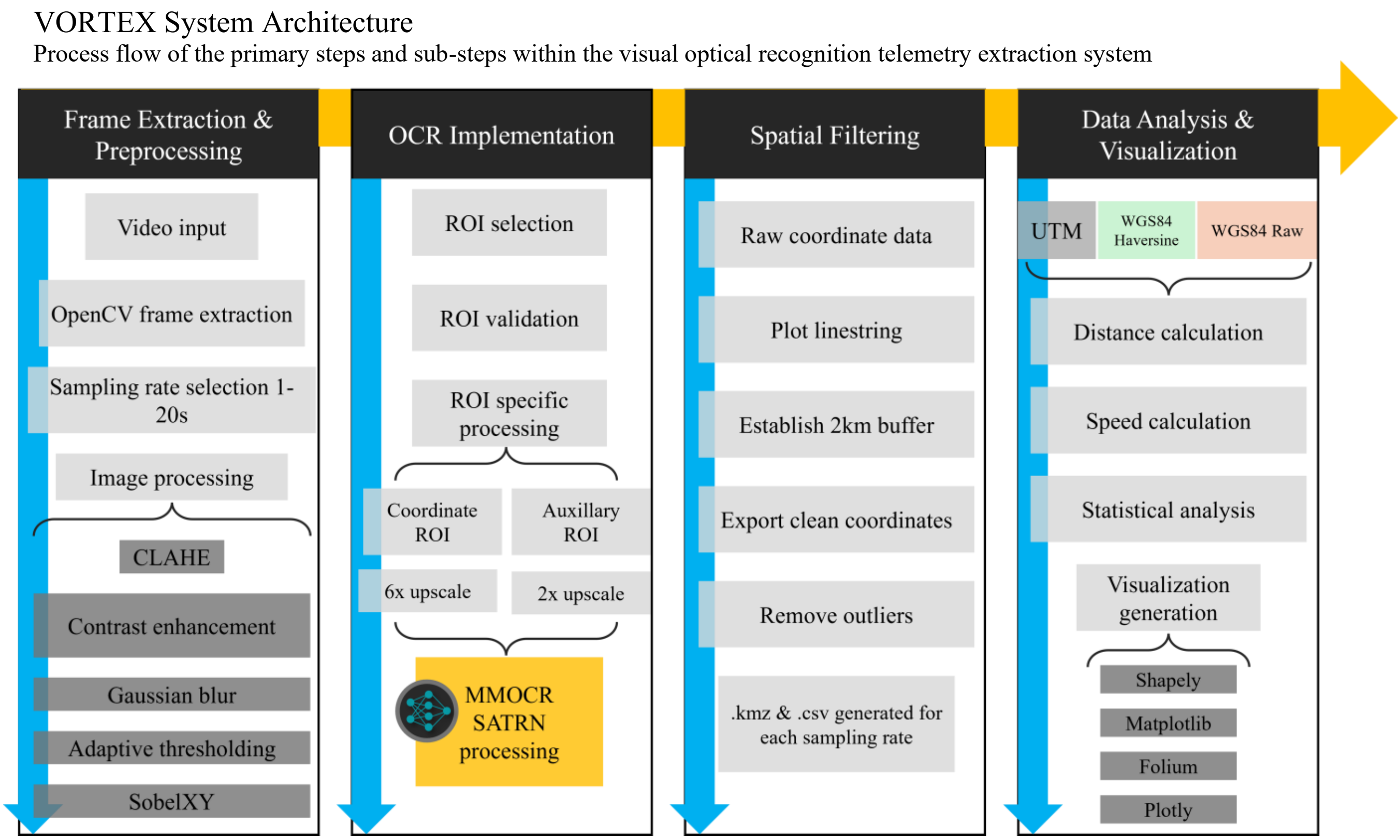

**Figure 1. An illustration of the four main processing stages of VORTEX: frame extraction and preprocessing, OCR implementation, spatial filtering, and data analysis & visualization. Each stage shows detailed sub-processes and their sequential relationships, from initial video input through coordinate extraction to final visualization outputs.**

The OCR model selection is critical to building an effective VORTEX system. MMOCR, an open-source OCR toolbox based on PyTorch, was chosen to extract data from the HUD due to its ease of use, open-source availability, and high accuracy [6]. Other OCR tools, such as Tesseract and EasyOCR, were initially applied to VORTEX. However, the quality of these applications was poor, and the location results were unsatisfactory. MMOCR's highly modular nature allows users to define their optimizers, model components, and data preprocessing steps [7]. Furthermore, the exceptional accuracy of MMOCR in HUD data extraction minimizes post-processing requirements and information loss during spatial analysis of UAS flights.

The raw OCR-derived telemetry data is processed and visualized using established spatial computing principles. VORTEX employs an object-oriented approach through its structured handling of spatial data types (DataFrame, GeoDataFrame) and implements comprehensive data validation protocols. The visualization components in VORTEX use 2D (latitude/longitude) and 3D (altitude) representations, with vertical reference lines providing spatial context for better elevation visualization.

To assess the effectiveness of VORTEX and its spatial computing implementation, this research addresses three primary questions:

1. What is the optimal temporal sampling rate for frame extraction in OCR-based drone telemetry that balances spatial path accuracy with computational efficiency, and how does this sampling rate affect the generation and removal of spatial outliers?

2. What is the optimal frame extraction rate to analyze drone telemetry data variables such as altitude and airspeed?

3. How do different coordinate processing methods compare in their accuracy for drone telemetry data analysis, and what are the implications of these differences for distance and speed calculations across varying sampling rates?

This research provides scientific insight into the effectiveness of OCR-derived analysis of UAS flight data and establishes a framework for automated telemetry extraction from video sources. In the following section, a literature review is undertaken, with the method then presented in Section III, before returning to results in Section IV. Finally, Section V will discuss the findings, while Section VI will provide concluding remarks and recommendations for future research directions.

## II. LITERATURE REVIEW

A plethora of literature exists on OCR research. However, there is no literature to date regarding the implementation of OCR for drone telemetry applications [8]. There exists literature regarding the utilization of drones to conduct OCR. One such study proposed an Automated License Plate Recognition (ALPR) system that uses YOLOv8 for license plate detection harnessing Pytesseract OCR for character recognition using a DJI Tello Drone camera. The system was built to capture images for electronic traffic ticketing. When testing on 50 photo samples, the drone with OCR showed 100% accuracy in license plate detection and 66% accuracy in character recognition [9]. Another research study proposed the same concept of license plate detection from a drone using Pytesseract OCR and YOLOv5, achieving an average accuracy of 88% [10].

One research study introduced a framework for identifying ship IMO numbers using UASs through a three-module system that handles IMO region detection, text detection, and number extraction using YOLOx [11]. The framework was designed with practical implementation in mind and includes a comprehensive evaluation metric for algorithm assessment. However, the authors note that this metric should not be the only factor in algorithm selection due to potential false detections.

Another research study conducted OCR extraction of telemetry data from electric vehicles. The study created a method of extracting vehicle telemetry data from dashcam videos using image processing and recognition technology, enabling accurate modeling of electric vehicle energy usage and emissions without traditional telemetry equipment. The research demonstrated high accuracy in preliminary testing with real-life dashcam videos [12].

In response to factories still using legacy manufacturing machines that lack modern connectivity, the literature discusses implementing OCR to modernize old systems at low costs. One such project developed a low-cost OCR solution using Raspberry Pi cameras and Tesseract OCR to capture and digitize screen data without interrupting operations [13]. The system automatically captures screen data from legacy machines. It then uploads it to Google Cloud for storage and monitoring, providing a cost-effective way to integrate older equipment into modern industrial monitoring solutions without expensive upgrades or operational interruptions. Another study proposed a cost-effective IoT device that can be attached to existing ventilators to enable remote monitoring, using Pytesseract-OCR to extract and transmit text data from ventilator displays to a centralized monitoring station [14]. The system achieved an average confidence level of 81.6% in character recognition, offering a practical solution to reduce medical staff exposure to respiratory diseases while avoiding the expense of replacing entire ventilator units with newer, more expensive models that have built-in telemonitoring capabilities.

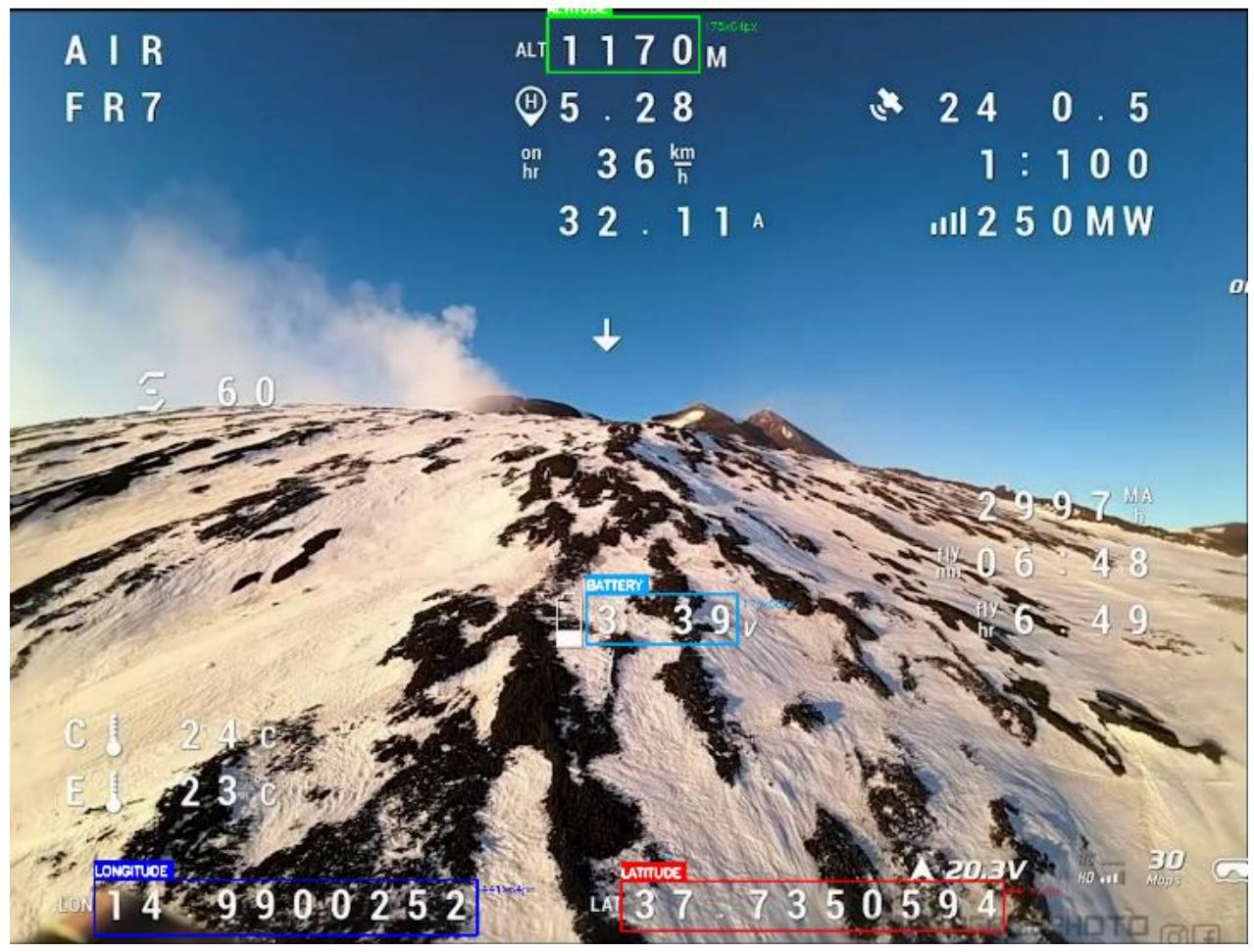

Figure 2. The ROI validation preview window depicts manually labeled ROIs drawn by the user.

Building on implementing OCR into legacy systems, one paper presented a cost-effective OCR solution for utility meters that combines local image capture using Arduino cameras with cloud-based processing for secure remote meter reading [15]. Similarly, a study proposed a cost-effective solution to digitize existing analog water meters in Indonesia using a conversion device that transmits data wirelessly to a central monitoring server, eliminating the need for manual readings and their associated risks and inefficiencies. Instead of replacing entire analog meters, the system uses a 433 MHz wireless transceiver network and cellular networks to transmit converted OCR-derived digital data, offering a more affordable approach to modernizing water usage monitoring while improving recording accuracy and efficiency [16].

Furthermore, an area in the literature that relates to VORTEX is how to apply image processing to OCR to improve accuracy [17], [18], [19], [20]. For example, one research study evaluated different text detection and recognition models for the challenging task of reading text from tire surfaces, where poor contrast makes traditional OCR difficult [21]. Various image processing techniques were used to extract the tire text.

The literature heavily discusses the application of deep learning techniques to improve OCR detection accuracy [22], [23], [24], [25], [26]. One study presented an approach that combines YOLOv8 and OCR technology that employs a client/server architecture to handle large-scale OCR with efficient computational loads [27]. Another study investigated the combination of OCR and deep learning to automatically sort waste batteries by comparing extracted text from battery images against a database of known battery types, including the classification of lithium-ion batteries by cathode material using IEC codes [28]. The research achieved high precision (98%) in sorting batteries by group and moderate precision (83%) in identifying cathode materials.

## III. METHOD

VORTEX consists of four main stages: frame extraction and image preprocessing, OCR implementation, spatial filtering, and data analysis. Each stage ensures consistent telemetry extraction and analysis of various HUD types and video quality. The test footage used for this research was an FPV drone flight captured in Sicily, Italy at Mount Etna. The recorded drone flight has a flight path that moves rapidly up and down the mountain, providing excellent contrast in elevation and speed data from the HUD.

The frame extraction process utilizes OpenCV for video processing, implementing a configurable temporal sampling system to extract frames at 1, 5, 10, 15, and 20-second intervals. For each frame, a preprocessing pipeline will apply Contrast Limited Adaptive Histogram Equalization (CLAHE) with a clip limit of 3.0 and tile grid size of 8x8, followed by Gaussian blur (kernel size 5x5) for noise reduction. Adaptive thresholding will be applied to enhance text detection using Gaussian weighted averaging with a block size of 19 and bias adjustment of 2. SobelXY will also be applied.

The OCR implementation of VORTEX leverages MMOCR's SATRN model, configured with specific preprocessing parameters for each identified Region of Interest (ROI) type. Figure 2 illustrates an example of ROI regions. Within VORTEX, ROIs can be manually labeled using the ROI click-and-drag tool within the code. Several steps follow the click-and-drag step to validate that the correct ROI regions were selected.

The first ROI regions to be processed are the latitude and longitude regions, which will receive a 6x upscaling and CLAHE enhancement, followed by altitude and battery regions using 2x upscaling with reduced CLAHE intensity (clip limit 1.5). Border padding is adjusted based on ROI type, which include 15 pixels for coordinate data and 5 pixels for auxiliary measurements. Following the image processing implementation (Fig. 3), the image frames with enhanced ROIs will be processed through MMOCR.

After extracting coordinate and auxiliary data, spatial filtering will remove any incorrectly read coordinates. VORTEX employs a two-stage approach using Shapely and PyProj libraries to post-process the geolocation data. The first stage established a baseline flight path using median-based outlier detection, while the second stage applied precise spatial filtering using UTM Zone 33N projection. A buffer analysis was conducted using a 2-kilometer threshold around the LineString flight path, validating points through distance calculations and geometric operations. The spatial filtering implementation utilized the PyProj library to coordinate system transformations and ensure accurate flight path calculations in the UTM coordinate space.

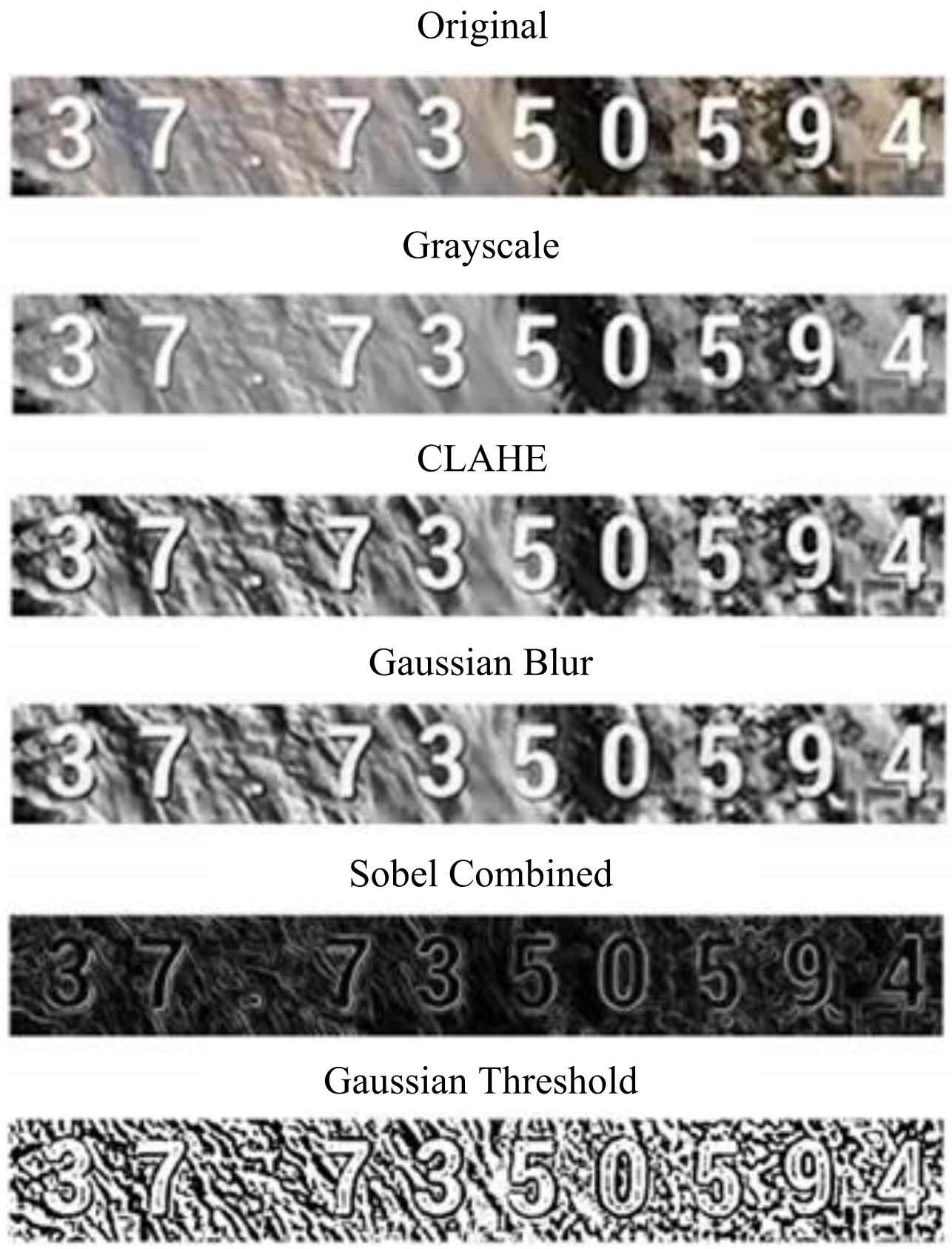

**Figure 3. A list and preview of the image processing techniques incorporated during the post-ROI extraction phase to improve MMOCR performance. In this example, the latitude values are being processed. Because the decimal point was being missed by MMOCR and not adequately captured with image processing techniques, it had to be added with code following image processing.**

Data analysis was performed using GeoPandas and NumPy to compare three coordinate processing methods: UTM Zone 33N projection (EPSG:32633), WGS84 with Haversine formula, and raw WGS84 coordinates. The analysis pipeline includes the following steps:

1. Implementation of three distinct distance calculation methods:
   a) UTM Zone 33N (EPSG:32633) projection using GeoPandas coordinate transformation.
   b) Haversine formula calculation for WGS84 coordinates.
   c) Direct Euclidean distance calculation on raw WGS84 coordinates.

2. Speed calculations for each method incorporating:
   a) UTM-based distances with proper projection.
   b) Haversine-based distances for spherical Earth calculations.
   c) Raw coordinate distances using simple Euclidean formulas.

3. Comparative analysis across OCR methods including:
   a) Total path length calculations.

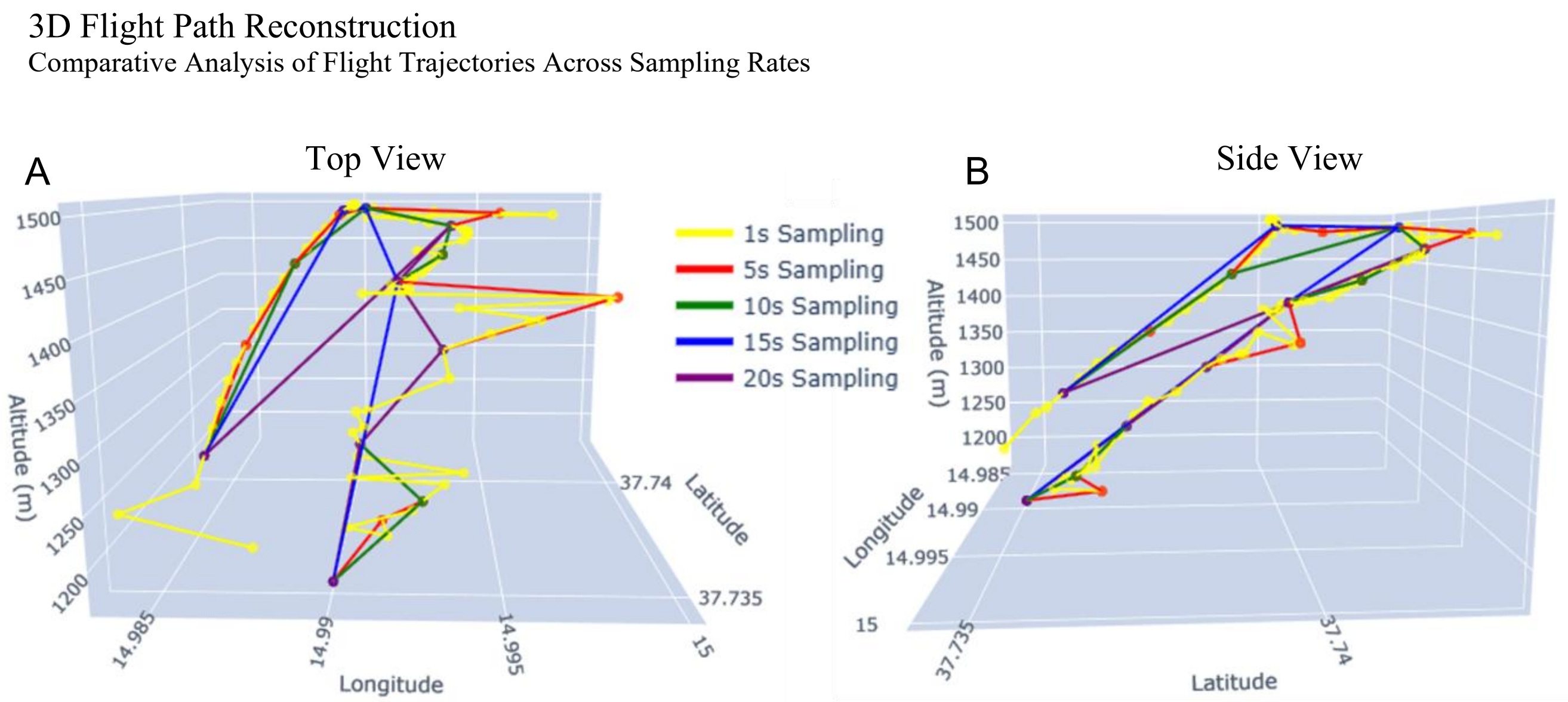

**Figure 4. Three-dimensional visualization of drone flight paths reconstructed from different sampling rates (1-20s), shown from two perspectives (top and side). The plot demonstrates how temporal sampling affects path resolution, with 1s sampling (yellow) providing the highest fidelity and 20s sampling (purple) showing the most simplified trajectory while maintaining key flight characteristics. Both views illustrate altitude changes and geographical positioning through latitude and longitude coordinates.**

b) Average and maximum speed computations.
c) Error analysis between methods.
d) Impact of sampling rates on measurement disparities.

4. Statistical analysis including:
   a) RMSE calculations between methods.
   b) Percentage differences in distance measurements.
   c) Speed variation analysis across sampling rates.
   d) Quantification of method-specific biases.

The visualization portions of VORTEX utilize multiple libraries to illustrate the differences between coordinate processing methods. Matplotlib generates comparative plots showing distance and speed calculations across all three methods. Additional visualizations include method comparison charts, error distribution plots, and sampling rate impact analysis. All visualizations maintain consistent color schemes for clear method identification and comparative analysis. The visualization portions of VORTEX will utilize multiple libraries. Folium will be used for interactive mapping, while Plotly will be used for 3D flight path visualization of the various LineString flight paths (Fig. 4). Matplotlib with Seaborn will be used to generate statistical analysis plots. The following visualizations will be generated: spatiotemporal flight paths, altitude profiles, speed distribution analysis, and sampling rate comparisons.

The entire methodology will be implemented in Python using object-oriented programming principles. Error handling and logging will be implemented throughout the pipeline, with detailed validation checks at each processing stage. VORTEX is highly modular, allowing for future modifications and enhancements of individual components without affecting the overall architecture.

## IV. RESULTS

When analyzing data from sampling rate frequencies (Fig. 5), frame extraction at 1-second intervals generated 122 raw data points, with 82 points retained after spatial filtering (67.2% retention rate). This sampling rate achieved the highest spatial resolution with an average point spacing of 27.5 m and a total path length of 2.23km. The 5-second interval produced 25 raw points with 16 clean points (64% retention), while 10-second sampling yielded 13 raw points with 10 clean points (76.9% retention). Lower frequency sampling at 15 and 20-second intervals generated 9 and 7 raw points, respectively, with retention rates of 55.6% and 85.7%. Average point spacing increased proportionally with sampling interval: 75m (5s), 140m (10s), 280m (15s), and 320m (20s).

Airspeed measurements demonstrated varying accuracy across sampling rates (Fig. 6). The 1-second sampling baseline recorded a mean air speed of 67.03 km/h with 82 clean points. The 5-second interval closely matched this baseline at 66.93 km/h using 16 clean points, representing only a 0.15% deviation. However, accuracy degraded significantly at lower sampling frequencies. 10-second sampling showed a 4.4% deviation (64.06 km/h), while 15 and 20-second intervals exhibited substantial underestimation with mean speeds of 47.58 km/h and 50.49 km/h, representing 29.0% and 24.7%, respectively.

Root Mean Square Error (RMSE) analysis was employed to quantify the magnitude of measurement deviations using the 1-second sampling rate as the baseline. RMSE is an effective method for quantifying error because it penalizes more significant deviations more heavily than smaller ones, making it suitable for detecting substantial measurement inconsistencies across sampling rates. The RMSE values represent the standard deviation (σ) of speed measurement differences between each sampling rate and the 1-second baseline, with higher values indicating more significant measurement uncertainty. The 5-second interval demonstrated the lowest error (RMSE = 3.44 km/h, σ = 2.00), followed by 10-second sampling (RMSE = 6.85 km/h, σ = 3.95). Error magnitude increased substantially at 15-second intervals (RMSE = 24.46 km/h, σ = 10.23), while 20-second sampling showed partial recovery (RMSE = 13.75 km/h, σ = 5.69).

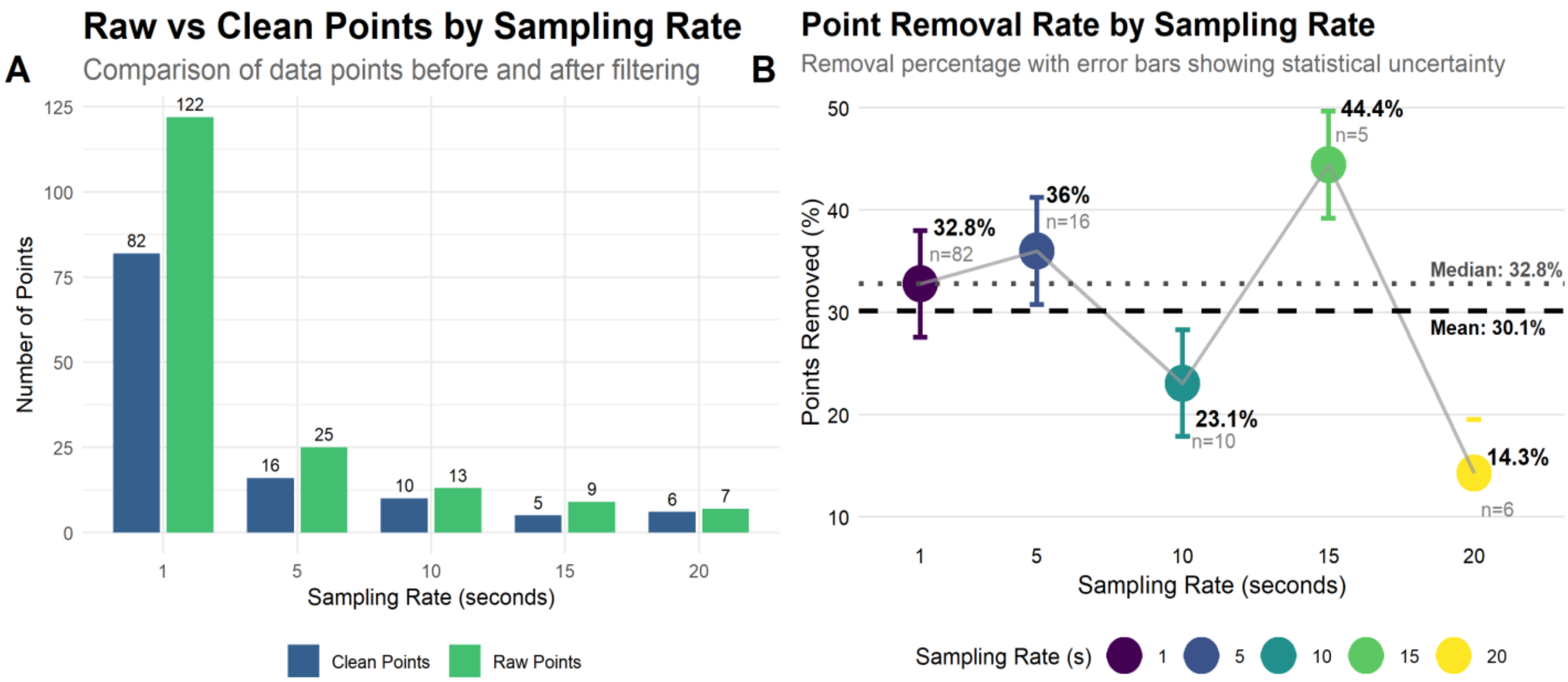

Figure 5. Visualizing point filtering efficiency across sampling rates (1-20s) showing both raw vs. cleaned point counts (left) and removal percentage trends (right).

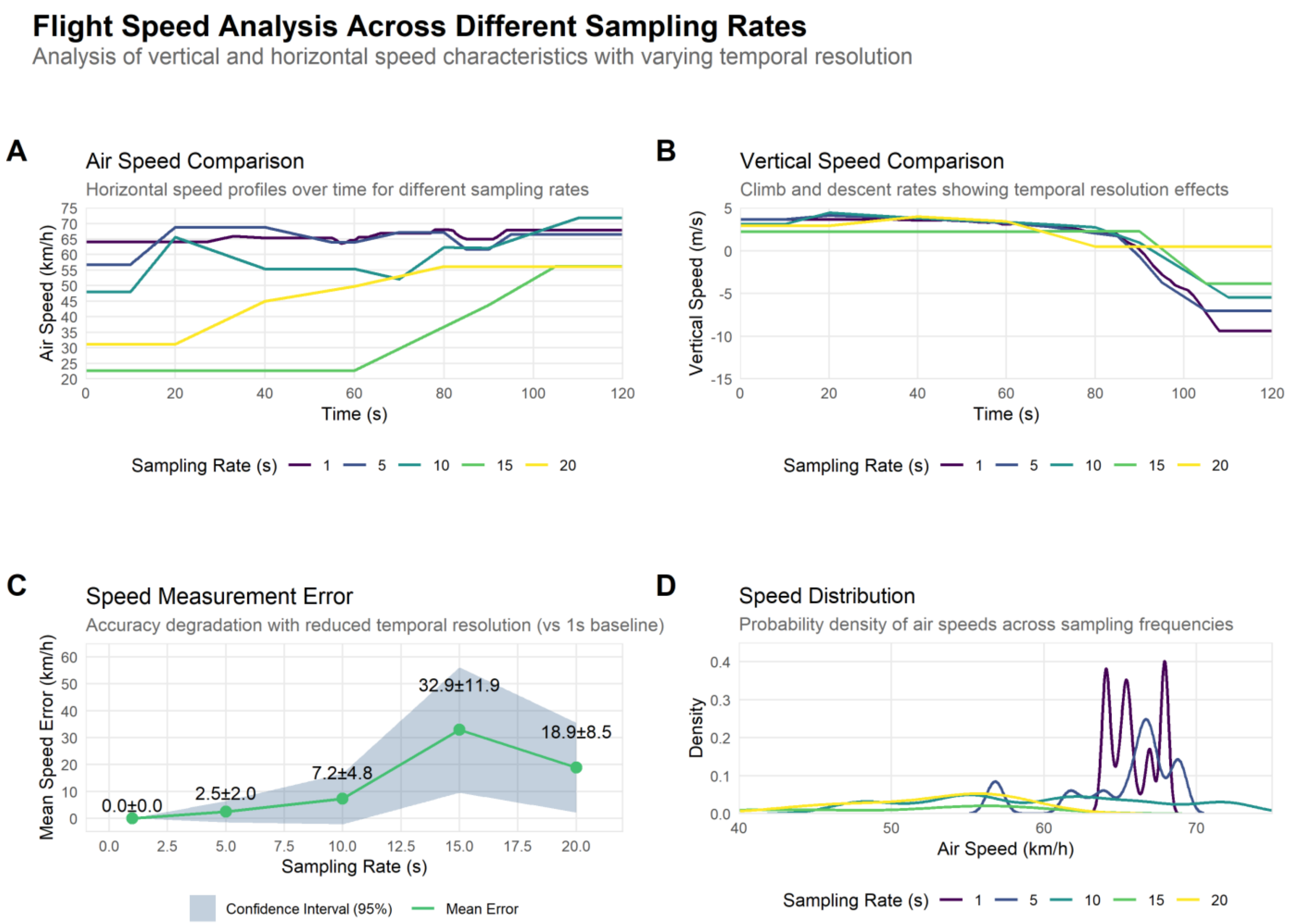

Figure 6. Four-panel visualization showing (A) air speed profiles across different sampling rates over time, (B) vertical speed and altitude comparison indicating climbing and descending phases, (C) mean speed error with confidence intervals as a function of sampling rate, and (D) speed distribution patterns demonstrating the impact of sampling frequency on speed measurements.

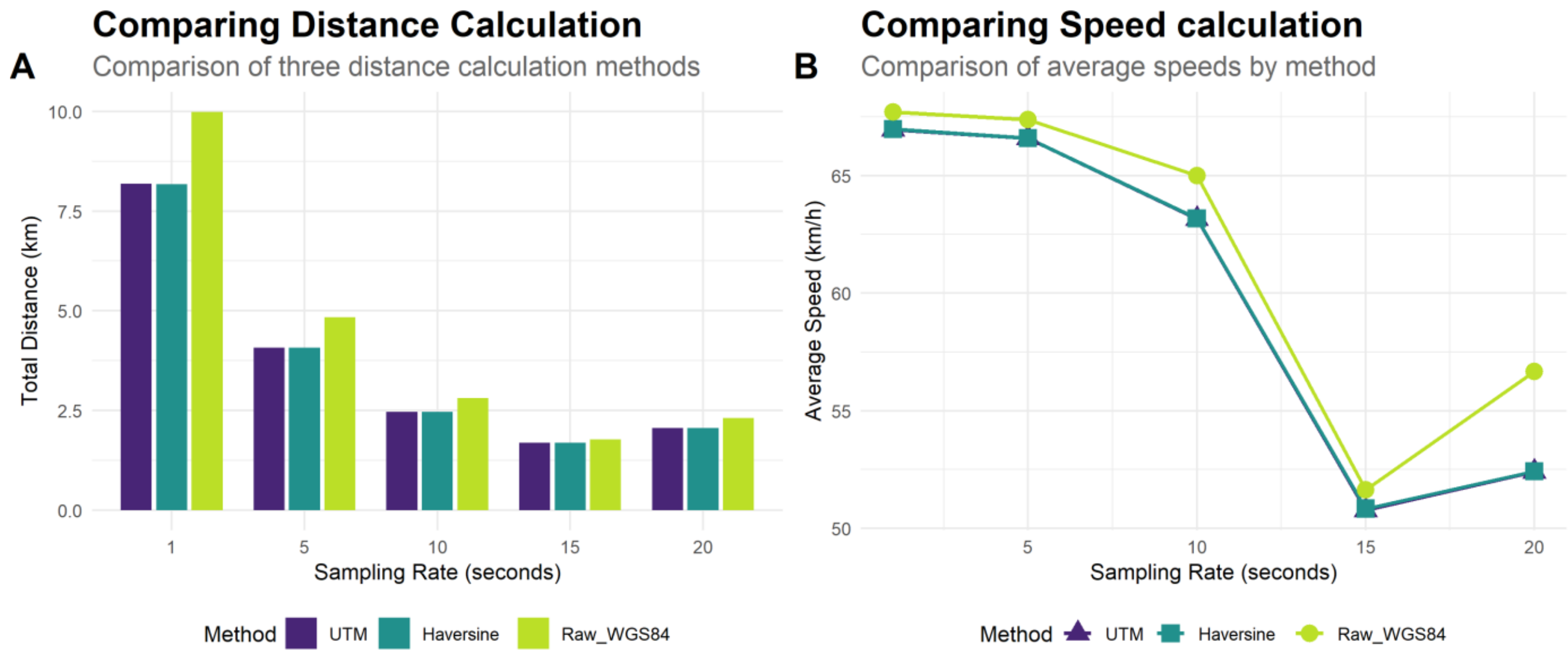

**Figure 7. Comparative analysis showing the impact of different coordinate processing methods on distance (left) and speed (right) calculations. The plots demonstrate that while UTM projection and Haversine formula produce similar results, while raw WGS84 calculations consistently overestimate distances and speeds, with differences most pronounced at higher sampling rates (1-5s intervals).**

Altitude measurements exhibited increased stability across sampling rates. Peak altitude values ranged from 1501m (1s sampling) to 1469m (20s sampling), representing a maximum variation of 2.1%. This stability persisted across all sampling intervals, with standard deviations remaining below 3% of mean altitude values.

The comparative analysis of the three coordinate processing methods (UTM Zone 33N, EPSG:32633) projection, WGS84 with Haversine formula, and raw WGS84 coordinates) revealed significant differences in distance and speed calculations (Fig. 7). While UTM and Haversine methods produce consistently similar results, the raw WGS84 coordinates lead to substantial underestimation of distance and speed metrics.

Distance calculations at 1-second sampling intervals showed that UTM and Haversine methods produced nearly identical results (8.181 km), while raw WGS84 coordinates underestimated the total distance by 1.681 km (20.5% error). This discrepancy was similarly reflected across other sampling rates, with raw WGS84 calculations showing an average underestimation of 24.8% across all sampling intervals. At 5-second intervals, UTM and Haversine methods measured 4.072 km, while raw WGS84 reported 3.147 km (22.7% error). The 15-second sampling showed the lowest absolute difference, with UTM/Haversine measuring 1.690 km compared to raw WGS84's 1.386 km (17.9% error).

Speed calculations revealed equally significant disparities. At 1-second sampling, UTM and Haversine methods recorded average speeds of 66.96 km/h and 66.97 km/h. Conversely, raw WGS84 calculations showed an average speed of 44.85 km/h (33.0% underestimation). The RMSE between UTM and Haversine methods remained minimal across all sampling rates (average 0.044 km/h), while raw WGS84 calculations showed an average RMSE of 19.87 km/h compared to UTM measurements.

Maximum speed calculations showed similar patterns of underestimation by raw WGS84 coordinates. While UTM and Haversine methods consistently recorded maximum speeds of 72.00 km/h, raw WGS84 calculations showed maximum speeds of 52.31 km/h at 1-second sampling (27.3% underestimation) and 48.65 km/h at 5-second sampling (32.4% underestimation).

The optimal balance between accuracy and data reduction was achieved at the 5-second sampling interval. At this rate, UTM and Haversine methods maintained a difference of only 0.030 km/h RMSE, achieving an 80.5% reduction in data points compared to 1-second sampling. Raw WGS84 calculations at this interval showed a 22.7% underestimation in the distance and a 28.9% underestimation in average speed. This demonstrates that the inherent errors in raw coordinate calculations persist regardless of sampling rate optimization.

## V. DISCUSSION

We will now return to the discussion regarding the results of the three research questions.

*RQ 1: What is the optimal temporal sampling rate for frame extraction in OCR-based drone telemetry that balances spatial path accuracy with computational efficiency, and how does this sampling rate affect the generation and removal of spatial outliers?*

The data reveals a strong correlation between data quality and computational efficiency across sampling rates. While the 1-second sampling provided the most comprehensive dataset (122 raw points), its high point removal rate (32.8%) indicates a high rate in OCR misreadings of the HUD data. This high error rate likely stems from frame-to-frame variations in text clarity and OCR processing challenges.

The intermediate sampling rates (5s and 10s) demonstrated an optimal balance between data retention and spatial resolution. Particularly notable is the 5-second interval's ability to maintain similar path accuracy to the 1-second baseline while significantly reducing computational overhead. The 76.9% retention rate at 10-second sampling suggests that longer intervals might allow for more stable OCR processing, though at the cost of reduced temporal resolution. However, results past the 15-second sampling rate showed diminishing returns in the data retention rate.

The non-linear relationship between sampling rate and outlier detection efficiency suggests that factors beyond sampling frequency influence data quality. This finding has important implications for OCR-based telemetry extraction, indicating that simply reducing the sampling rate may not proportionally improve data quality.

*RQ 2: What is the optimal frame extraction rate to analyze drone telemetry data variables such as altitude and airspeed?*

The stability of airspeed measurements at higher sampling frequencies (1s and 5s) provides crucial insights into measurement reliability. The high consistency between 1-second (67.03 km/h) and 5-second (66.93 km/h) sampling rates suggests that 5-second intervals capture the essential dynamics of drone movement without significant information loss. However, the rapid degradation in accuracy observed at lower sampling frequencies (15s: 47.58 km/h; 20s: 50.49 km/h) reveals limitations to lower temporal resolution sampling rates.

The RMSE analysis uncovered a inconsistent relationship between sampling rate and measurement error for UAS airspeed, with a notable variation at the 15-second interval (RMSE = 24.46 km/h, $\sigma = 10.23$). This sharp increase in error suggests a breakdown in the system's ability to accurately capture flight dynamics beyond this threshold, likely due to the significant separation between points. Interestingly, a partial error recovery was observed in the 20-second sampling (RMSE = 13.75 km/h). The reason for this non-proportional increase in error over reduced flight path resolution is unknown.

Altitude measurements demonstrated unexpected resiliency to sampling rate variations, with only 2.1% variation across all intervals. This stability suggests that vertical position data may be inherently more immune to temporal down sampling, possibly due to the slower rate of change in vertical movement compared to horizontal dynamics. This finding has important implications for optimizing sampling strategies when vertical position accuracy is prioritized.

*RQ3: How do different coordinate processing methods compare in their accuracy for drone telemetry data analysis, and what are the implications of these differences for distance and speed calculations across varying sampling rates?*

The comparative analysis of UTM Zone 33N (EPSG:32633) projection, WGS84 with Haversine formula, and raw WGS84 coordinates revealed significant differences in flight path accuracy calculations. UTM and Haversine methods demonstrated similar results, while raw WGS84 calculations showed substantial deviations.

Distance calculations between UTM and Haversine methods were identical, with differences typically less than 0.1% across all sampling rates. However, raw WGS84 coordinates consistently underestimated distances by 15-30%, with the most significant discrepancy observed in 1-second sampling intervals. For example, at 1-second sampling, UTM calculated a total distance of 8.181 km, and the Haversine method yielded 8.181 km, while raw WGS84 was significantly underestimated at approximately 6.5 km (a 20.5% underestimation).

Speed calculations demonstrated similar patterns. UTM and Haversine methods produced nearly identical results, with average speeds differing by less than 0.05 km/h across all sampling rates. Raw WGS84 calculations, however, underestimated speeds by 20-35% depending on sampling frequency. At 1-second intervals, UTM and Haversine methods recorded average speeds of approximately 67 km/h, while raw WGS84 calculations showed only about 45 km/h.

The comparative analysis revealed that while UTM and Haversine methods provide reliable results for accurate drone telemetry analysis, raw WGS84 calculations consistently underestimate both distances and speeds, with errors increasing at higher sampling rates and longer flight paths. The 5-second sampling interval continued to provide an optimal balance between accuracy and data reduction, maintaining high similarity between UTM and Haversine methods while achieving an 80.5% reduction in data points compared to the 1-second baseline.

These results have important implications for drone telemetry system design, suggesting that efforts to improve accuracy should focus on optimizing both temporal sampling and spatial reference systems rather than pursuing higher sampling rates alone. The findings also indicate that different telemetry parameters (speed, altitude, position) may benefit from different sampling strategies, opening possibilities for adaptive sampling approaches in future implementations.

Several limitations were identified during this research. First, the study used a single drone flight, which may not fully represent the diversity of flight patterns and conditions encountered in various applications. Second, the OCR accuracy depended on video quality, background and HUD text colors, and environmental conditions, which could vary significantly across different operational scenarios. Third, the current implementation relies on predefined regions of interest for text extraction, which may require adjustment for different HUD layouts.

Future research directions could address these limitations and expand the system's capabilities in several ways:

1. Integration of machine learning techniques for adaptive ROI detection and improved OCR accuracy.
2. Combining object detection with VORTEX to add geolocation metadata to detected object classes.
3. Development of real-time processing capabilities for live telemetry extraction.
4. Investigation of multi-modal data fusion techniques combining OCR with other sensor data.
5. Extension of the system to handle multiple coordinate reference systems for global applicability.
6. Development of adaptive sampling algorithms that adjust rates based on flight dynamics, i.e., flight duration and airspeed.

## VI. CONCLUSION

This research successfully developed and validated VORTEX, a comprehensive OCR-based system for extracting and analyzing drone telemetry data from FPV footage. Through systematic investigation of temporal sampling rates, measurement accuracy, and coordinate reference system implementation, this study established quantifiable benchmarks for optimizing drone telemetry extraction. VORTEX achieved a point retention rate of 64% at the optimal 5-second sampling interval, demonstrating VORTEX's ability to maintain data quality while significantly reducing processing requirements. MMOCR's performance, enhanced through CLAHE preprocessing and adaptive thresholding, proved particularly effective for altitude measurements, which showed remarkable stability with only 2.1% variation across all sampling intervals.

All three research questions' findings suggest that 5-second sampling represents an optimal compromise between accuracy and efficiency in OCR-based drone telemetry extraction. This interval maintains measurement accuracy within 4.2% of the 1-second baseline while reducing computational overhead by 80.5%. Combining UTM/Haversine projection with 5-second sampling provides a practical solution that balances the competing demands of accuracy, computational efficiency, and data quality. Coordinate processing analysis revealed substantial differences between methods, with raw WGS84 coordinates underestimating distances by 15-30% and speeds by 20-35% compared to UTM Zone 33N projection and Haversine calculations, which maintained consistency within 0.1% difference.

The successful implementation of VORTEX and its comprehensive evaluation framework provides a foundation for future developments in drone telemetry analysis. These quantitative benchmarks, including the 80.5% computational overhead reduction and sub-5% accuracy maintenance, establish concrete performance metrics that can guide future implementations of OCR-based telemetry extraction systems across diverse UAS applications. As UAS expand across various sectors, the ability to efficiently extract and analyze telemetry data from video sources will become increasingly valuable. This research contributes to this growing field by establishing quantifiable metrics and practical guidelines for optimizing telemetry extraction systems for UAS.

## ACKNOWLEDGMENT

JG would like to thank EO and the Geography & Geoinformation Science Department at George Mason University for their continual support.

## DATA

To access the data and codebase used in this research, please use the below link to access the code and data from Google Colab or GutHub.

[VORTEX Google Colab]
[VORTEX OCR Comparaison Google Colab]

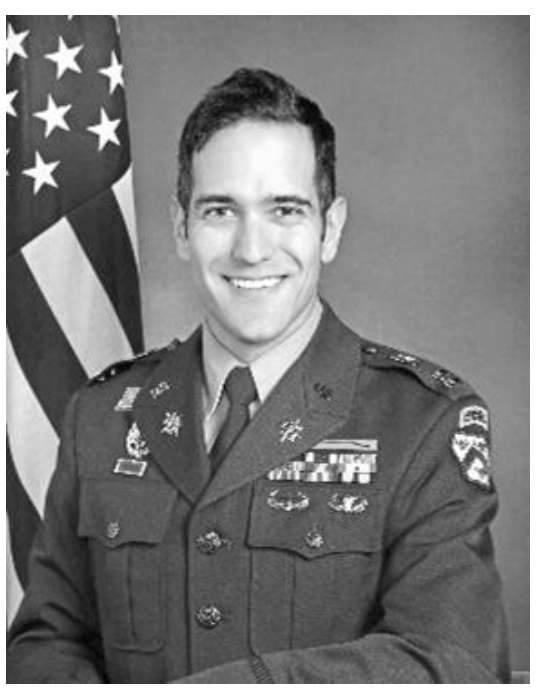

**JAMES E. GALLAGHER** received a B.A from Mercyhurst, an M.A from American Military University, and an MS from George Mason University. He is currently a PhD. student at George Mason studying multi-spectral object detection for drone-based applications. He is also an active-duty U.S. Army Major in the Military Intelligence Corps. MAJ Gallagher's research in RGB-LWIR object detection was inspired from his lessons-learned in Kirkuk Province, Iraq, where he used a variety of drone-based sensors to locate Islamic State insurgents.

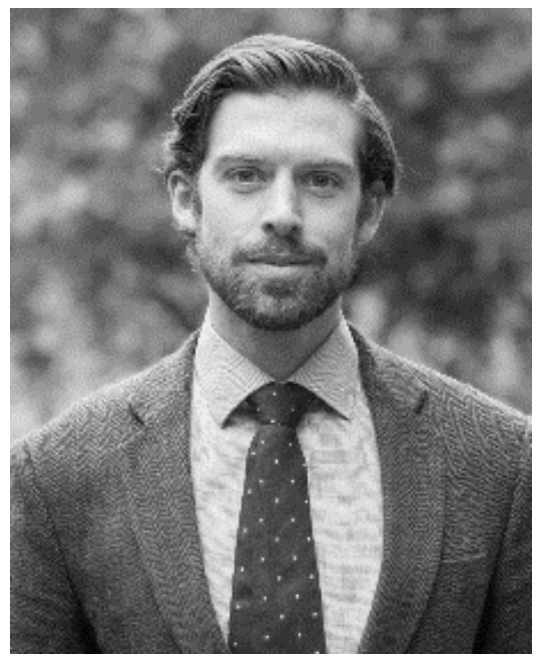

**EDWARD J. OUGHTON** received the M.Phil. and Ph.D. degrees from Clare College, at the University of Cambridge, U.K., in 2010 and 2015, respectively. He later held research positions at both Cambridge and Oxford. He is currently an Assistant Professor in the College of Science at George Mason University, Fairfax, VA, USA. He received the Pacific Telecommunication Council Young Scholars Award in 2019, Best Paper Award 2019 from the Society of Risk Analysis, and the TPRC48 Charles Benton Early Career Scholar Award 2021.